\documentclass[runningheads,a4paper]{llncs}
\usepackage{amsmath}
\usepackage{amsfonts}
\usepackage{amssymb}
\usepackage{graphicx}
\usepackage{enumerate}
\usepackage[capitalise]{cleveref}
\crefname{section}{Sect.}{Sect.}
\crefname{equation}{}{}
\usepackage{bm}
\usepackage{url}
\usepackage{tikz}
\usetikzlibrary{positioning}
\usetikzlibrary{shapes.geometric}
\usepackage{circuitikz}
\usepackage{algorithm}
\usepackage{algpseudocode}
\algnewcommand\algorithmicinput{\textbf{Input:}}
\algnewcommand\Input{\item[\algorithmicinput]}
\algnewcommand\algorithmicoutput{\textbf{Output:}}
\algnewcommand\Output{\item[\algorithmicoutput]}
\algnewcommand\algorithmicforeach{\textbf{for each}}
\algdef{S}[FOR]{ForEach}[1]{\algorithmicforeach\ #1\ \algorithmicdo}

\makeatletter
\AtBeginDocument{%
  \def\doi#1{\url{https://doi.org/#1}}}
\makeatother

\newcommand{\ct}[1]{\cos\theta_{#1}}
\newcommand{\st}[1]{\sin\theta_{#1}}
\newcommand{\cd}[1]{\cos\delta_{#1}}
\newcommand{\sd}[1]{\sin\delta_{#1}}
\newcommand{\ca}[1]{\cos\alpha_{#1}}
\newcommand{\sa}[1]{\sin\alpha_{#1}}
\newcommand{\cbe}[1]{\cos\beta_{#1}}
\newcommand{\sbe}[1]{\sin\beta_{#1}}
\newcommand{\cga}[1]{\cos\gamma_{#1}}
\newcommand{\sga}[1]{\sin\gamma_{#1}}
\newcommand{\origin}[1]{\mathcal{O}_{#1}}

\newcommand{\Vector}[4]{
  \begin{pmatrix}
    #1 \\ #2 \\ #3 \\	#4
  \end{pmatrix}
}

\begin{document}
\title{
  Inverse Kinematics for a 6-Degree-of-Freedom Robot Manipulator Using Comprehensive Gr\"obner Systems
}
\titlerunning{Inverse Kinematics for a 6-Degree-of-Freedom Robot Manipulator}
\author{
    Takumu Okazaki\inst{1}\and
    Akira Terui\inst{1}\orcidID{0000-0003-0846-3643} \and
    Masahiko Mikawa\inst{1}\orcidID{0000-0002-2193-3198}
}
\institute{%
University of Tsukuba, Tsukuba, Japan \\
\email{terui@math.tsukuba.ac.jp}\\
\email{mikawa@slis.tsukuba.ac.jp}\\
\url{https://researchmap.jp/aterui}
}

\maketitle

\begin{abstract}
  We propose an effective method for solving the inverse kinematic problem
  of a specific model of 6-degree-of-freedom (6-DOF) robot manipulator using computer algebra.
  It is known that when the rotation axes of \emph{three} consecutive rotational joints of a manipulator intersect at a single point, the inverse kinematics problem can be divided into
  determining position and orientation.
  We extend this method to more general manipulators in which the rotational axes of
  \emph{two} consecutive joints intersect.
  This extension broadens the class of 6-DOF manipulators for which the inverse kinematics problem can be solved, and is expected to enable more efficient solutions.
  The inverse kinematic problem is solved using the Comprehensive Gr\"obner System (CGS) with
  joint parameters of the robot appearing as parameters in the coefficients
  to prevent repetitive calculations of the Gr\"obner bases.
  The effectiveness of the proposed method is shown by experiments.
\end{abstract}

\section{Introduction}
\label{sec:introduction}

\subsection{Problem Statements and Related Work}
\label{sec:inverse-kinematics}

This paper discusses the motion planning of a 6-Degree-of-Freedom (DOF) robot manipulator.
A manipulator is a robot resembling a human hand, consisting of
\emph{links} that function as a human arm and \emph{joints} as human joints
(see \cref{fig:mycobot-block-diagram}).
Each link is connected to the others by a joint. The first link is connected to the ground, and the last link, called the \emph{end-effector}, contains the hand, which can be moved freely.
In this paper, we discuss targeting a manipulator called ``myCobot 280'' \cite{mycobot-280}
(hereafter referred to as ``myCobot'').
It has six revolute joints, meaning each joint can rotate around its axis.
Thus, myCobot has 6-DOF.
As its manufacturer describes \cite{mycobot-280}, myCobot has been widely sold
for ``research and education, science and technology applications, and commercial exhibitions.''

In this paper, we deal with the inverse kinematic problem of the manipulator.
The inverse kinematic problem is a problem in finding the joint angles of the robot manipulator corresponding to (or achieving) a given end-effector's position and orientation.
In this study, we focus on two main challenges. The first is solving the inverse kinematics problem for a 6-degree-of-freedom (6-DOF) manipulator. The second is improving the efficiency of solving the inverse kinematics problem using Gr\"obner basis computation.

For the inverse kinematic problem of 6-DOF robot manipulators, several methods have been proposed,
including computer algebra \cite{hus-man2007,man-can1994,ni-wu2018}.
However, solving the inverse kinematics problem for 6-DOF manipulators is often challenging. In particular, algebraic approaches tend to result in large and complex systems.
For example, Pieper \cite{pie1968} has reported that, after elimination of variables with algebraic methods, the degree of the resulting univariate equation exceeds $500{,}000$, making it difficult to solve the inverse kinematics problem of a 6-DOF manipulator using algebraic methods in general.

Therefore, previously proposed methods solve the problem by appropriately dividing it into smaller sub-problems, thereby reducing the size of each problem and solving them with a realistic amount of computation \cite{hus-man2007,pie1968}.
Generally, a 6-DOF manipulator cannot independently determine the orientation and position of the end-effector.
Thus, it is difficult to solve the general inverse kinematics problem analytically.
\emph{However, if the manipulator includes a spherical joint with 3-dimensional rotation, the orientation and position of the end-effector can be determined independently.}
Then, the inverse kinematics problem can be separated into a 3-dimensional problem for orientation and a 3-dimensional problem for position.
Furthermore, Pieper \cite{pie1968} has demonstrated that if the rotational axes of three consecutive rotational joints intersect at a single point, the motion of these three joints is kinematically equivalent to a spherical joint. He has proposed a solution to the inverse kinematics problem, focusing on this intersection point.
This method was remarkable in solving the inverse kinematics problem for 6-DOF manipulators; however, it also has some limitations. One major drawback is that it cannot be applied unless the rotational axes of three consecutive joints intersect at a single point.

As for Gr\"obner basis computation, methods by reducing the inverse kinematic problem to a system of polynomial equations and using the Gr\"obner basis computation have been proposed for several types of robots, including manipulators \cite{fau-mer-rou2006,kal-kal1993,ni-wu2018,ric-sch-ces2021,uch-mcp2011,uch-mcp2012}.
Compared to numerical methods,
the use of the Gr\"obner basis computation has advantages, such that it can search for solutions
globally
so that the number of solutions of the inverse kinematics problem that can be obtained before the actual motion of the manipulator.

On the other hand, a disadvantage of the Gr\"obner basis computation is the relatively high computational cost compared to numerical methods, especially in the case of trajectory planning.
The trajectory planning problem of the manipulator is to determine a sequence of joint configurations that the manipulator should follow to move the end-effector along a specified trajectory.
In solving the trajectory planning problem, finitely many points on the trajectory along the path of the end-effector are given, and then we solve the inverse kinematic problem at each point.
When solving a trajectory planning problem,
the computational cost of repeatedly calculating the Gr\"obner basis at each point of the trajectory may become a significant issue.

\subsection{Our Previous Approach}
\label{sec:previous-approach}

We have proposed the following methods for the 3-DOF manipulator's inverse kinematic and trajectory planning problems.
Our first work \cite{ota-ter-mik2021} consists of the following:
1) For the inverse kinematic problem,
by using the Comprehensive Gr\"obner System (CGS) \cite{suz-sat2006}
and the real Quantifier Elimination (QE) based on the CGS \cite{fuk-iwa-sat2015},
called the CGS-QE method,
we have proposed a method to certify the existence of a solution to the inverse kinematic problem;
2) using this method for trajectory planning, we have proposed an efficient method that prevents the repeated calculation of the Gr\"obner basis.
In a follow-up paper \cite{yos-ter-mik2023},
by using the CGS-QE method, we have proposed a method to certify the existence of a solution to the inverse kinematic problem for the whole trajectory given as a line segment.
\subsection{Contributions}
\label{sec:contributions}

Based on the above results, we propose the following methods for the inverse kinematic problem
of a 6-DOF manipulator (myCobot).


We further extend Pieper’s method and propose a solution for the case where the rotational axes of two consecutive joints intersect at a single point, which is applicable to a wider variety of robots.
Our method enables us to globally and explicitly determine the solution to the inverse kinematics problem of myCobot, including the joint angles.

Furthermore, we use the CGS for solving a parametric system of polynomial equations,
which is important for avoiding recalculating the Gr\"obner basis in trajectory planning
 (see \Cref{sec:conclusion}).
 Note that in our experiments, the computation of the CGS has not been fully completed; however, in the experimental results for the inverse kinematics problems constructed by the authors from pre-existing randomly generated joint configurations, the inverse kinematics problems were successfully solved using the partially computed CGS.


\subsection{Organization of the Paper}
\label{sec:organization}

This paper is organized as follows.
In \Cref{sec:inverse-kinematics-mycobot}, we describe the inverse kinematic problem of myCobot.
In \Cref{sec:solving-inverse-kinematics}, we propose a method to solve the inverse kinematic problem of myCobot using the CGS, and present it in the form of an algorithm.
In \Cref{sec:experiments}, we show the effectiveness of the proposed method by experiments.
In \Cref{sec:conclusion}, we discuss conclusions and future work.

\section{The inverse kinematic problem of myCobot}
\label{sec:inverse-kinematics-mycobot}
\begin{figure}[t]
  \begin{minipage}[b]{0.3\textwidth}
    \centering
    \includegraphics[scale=0.35]{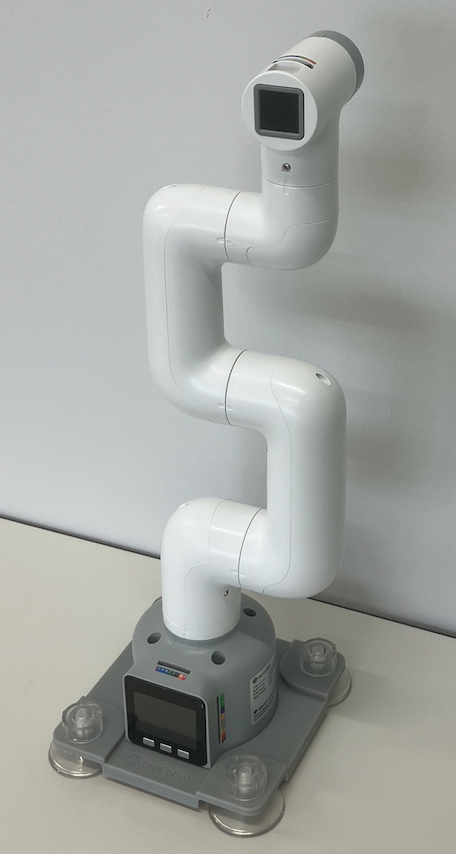}
    \caption{myCobot 280 \cite{mycobot-280}}
    \label{fig:mycobot}
  \end{minipage}
  \hfill
  \begin{minipage}[b]{0.65\textwidth}
  \centering
  \begin{tikzpicture}[scale=0.35]
  \draw (2,1)--(2,5.5);
  \draw (1.5,1.5) rectangle (2.5,3.5); 
  \draw (0.5,6)--(3.5,6)--(3.5,7)--(2,7)--(2,9.5);
  \draw (1,5.5) rectangle (3,6.5); 
  \draw (3.5,10)--(0.5,10)--(0.5,11)--(2,11)--(2,13.5);
  \draw (1,9.5) rectangle (3,10.5); 
  \draw (0.5,14)--(5.5,14);
  \draw (1,13.5) rectangle (3,14.5); 
  \draw (6,12.5)--(6,17.5);
  \draw (5.5,13) rectangle (6.5,15); 
  \draw (6,18) circle(0.5); 
  \draw (6-0.3535,18-0.3535)--(6+0.3535,18+0.3535); 
  \draw (6+0.3535,18-0.3535)--(6-0.3535,18+0.3535); 
  \draw (4,2.5)node{\small Joint 1};
  \draw (4,5)node{\small Joint 2};
  \draw (0.2,9)node{\small Joint 3};
  \draw (4,12.75)node{\small Joint 4};
  \draw (8.5,14)node{\small Joint 5};
  \draw (8.5,17.5)node{\small Joint 6};
  \draw (12,18.5)node{\small The end-effector (Joint 7)};
  \draw (1.5,4.5)node{ $d_1$};
  \draw (1.5,8)node{ $a_2$};
  \draw (1.5,12)node{ $a_3$};
  \draw (4.25,14.5)node{ $d_4$};
  \draw (5.5,16.25)node{ $d_5$};
  \draw (5,18)node{ $d_6$};
  \draw (1.9,1.2) to [out=180,in=90] (1.5,1);
  \draw (1.5,1) to [out=-90,in=180] (2,0.8);
  \draw (2,0.8) to [out=0,in=-90] (2.5,1);
  \draw[arrows = {- stealth}] (2.5,1) to [out=90,in=0] (2.1,1.2);
  \draw (3.1,1)node{ $\theta_1$};

  \draw (0.7,6.1) to [out=90,in=0] (0.5,6.5);
  \draw (0.5,6.5) to [out=180,in=90] (0.3,6);
  \draw (0.3,6) to [out=-90,in=180] (0.5,5.5);
  \draw[arrows = {- stealth}] (0.5,5.5) to [out=0,in=-90] (0.7,5.9);
  \draw (-0.2,6)node{ $\theta_2$};

  \draw (3.3,9.9) to [out=-90,in=180] (3.5,9.5);
  \draw (3.5,9.5) to [out=0,in=-90] (3.7,10);
  \draw (3.7,10) to [out=90,in=0] (3.5,10.5);
  \draw[arrows = {- stealth}] (3.5,10.5) to [out=180,in=90] (3.3,10.1);
  \draw (4.25,10)node{ $\theta_3$};

  \draw (0.7,14.1) to [out=90,in=0] (0.5,14.5);
  \draw (0.5,14.5) to [out=180,in=90] (0.3,14);
  \draw (0.3,14) to [out=-90,in=180] (0.5,13.5);
  \draw[arrows = {- stealth}] (0.5,13.5) to [out=0,in=-90] (0.7,13.9);
  \draw (-0.2,14)node{ $\theta_4$};

  \draw (5.9,12.7) to [out=180,in=90] (5.5,12.5);
  \draw (5.5,12.5) to [out=-90,in=180] (6,12.3);
  \draw (6,12.3) to [out=0,in=-90] (6.5,12.5);
  \draw[arrows = {- stealth}] (6.5,12.5) to [out=90,in=0] (6.1,12.7);
  \draw (6,11.5)node{ $\theta_5$};

  \draw (5.5,18.7) to [out=90,in=180] (6,19.2);
  \draw[arrows = {- stealth}] (6,19.2) to [out=0,in=90] (6.5,18.7);
  \draw (6,20)node{ $\theta_6$};
\end{tikzpicture}
\caption{myCobot with six rotational joints and five links}
\label{fig:mycobot-block-diagram}
\end{minipage}
\end{figure}

\subsection{Description of MyCobot}
\label{sec:description-mycobot}

myCobot 280 \cite{mycobot-280} (referred to as ``myCobot'', shown in \Cref{fig:mycobot}) is a 6-DOF robot manipulator with six rotational joints connected in series.
The arm length is 350 [mm], and it has a working radius of 280 [mm] centered at a height of approximately 130 [mm] from the ground.
myCobot can be controlled using programming languages such as Python, C++, and C\#. Additionally, it can be controlled using ROS (Robot Operating System) \cite{ROS2}, a standard control environment in robotics.
\Cref{fig:mycobot-block-diagram} shows a block diagram of myCobot.

\subsection{Description of the Transformation Matrix}
\label{sec:description-transformation-matrix}

In the following, we formulate the inverse kinematic problem of myCobot using the Denavit--Hartenberg convention
\cite{den-har1955} (referred to as ``D-H convention'').
First, we define the symbols necessary for defining the coordinate systems.
Let $\Sigma_i$ be the coordinate system w.r.t.\ Joint $i$, and ${}^ix,{}^iy$, and ${}^iz$ be
the $x$, $y$, and $z$-axis of $\Sigma_i$, respectively.
Let $\mathcal{O}_i$ be the origin of $\Sigma_i$, and
${}^{i-1}l_i$ be the common normal of the axes ${}^{i-1}z$ and ${}^iz$.

According to the D-H convention, for $i=1,\dots,7$, the coordinate system $\Sigma_i$ is defined
as a right-handed coordinate system that satisfies the following:
the origin $\mathcal{O}_i$ is placed at Joint $i$;
the axis ${}^iz$ is aligned with the rotation axis of Joint $i$, with the positive direction pointing towards Joint $i+1$;
the ${}^ix$ and ${}^iy$ axes follow the conventions of the 3D visualization tool RViz \cite{Rviz} in ROS.
Note that $\Sigma_1$ is treated as the global coordinate system.

Next, the transformation matrix from $\Sigma_{i+1}$ to $\Sigma_{i}$ is defined with the following parameters.
Note that, for each parameter, the unit of length is expressed in [mm], and the unit of angle is expressed in [rad].
Let $a_i$ be the length of the common normal ${}^{i}l_{i+1}$,
$\alpha_i$ the rotation angle around the ${}^{i+1}x$ axis between the ${}^iz$ and the ${}^{i+1}z$ axes,
$d_i$ the distance between the common normal ${}^{i}l_{i+1}$ and $\mathcal{O}_i$,
$\theta_i$ the rotation angle around the ${}^iz$ axis between the common normal ${}^{i}l_{i+1}$ and the ${}^ix$ axis.
Then, the transformation matrix ${}^iT_{i+1}$ from $\Sigma_{i+1}$ to $\Sigma_i$ is given by
\begin{equation}
  \label{eq:DHtransformation}
  \begin{split}
    {}^iT_{i+1}&=
    \begin{pmatrix}
      \ct{i} & -\st{i}\ca{i} & \st{i}\sa{i} & a_i \ct{i} \\
      \st{i} & \ct{i}\ca{i} & -\ct{i}\sa{i} & a_i\st{i} \\
      0 & \sa{i} & \ca{i} & d_i \\
      0 & 0 & 0 & 1
    \end{pmatrix}
    .
  \end{split}
\end{equation}

In the coordinate transformation of myCobot in RViz, in addition to the transformation in
\Cref{eq:DHtransformation}, we add $\delta_i$ as the rotation angle around the ${}^{i+1}z$ axis between the ${}^ix$ and the ${}^{i+1}x$ axes.
Thus, the transformation matrix $A_i$ from $\Sigma_{i+1}$ to $\Sigma_i$ with respect to the coordinate transformation in RViz is given as 
    \begin{multline}
      \label{eq:DH-RViz}
      \begin{split}
        A_i =
        \left(
        \begin{array}{cc}
          \cd{i}\ct{i}-\ca{i}\sd{i}\st{i} & -\ca{i}\cd{i}\st{i}-\sd{i}\ct{i} \\
          \ca{i}\sd{i}\ct{i}+\cd{i}\st{i} & \ca{i}\cd{i}\ct{i}-\sd{i}\st{i} \\
          \sa{i}\sd{i} & \sa{i}\cd{i} \\
          0 & 0
        \end{array}
        \right.
        \\
        \left.
          \begin{array}{cc}
            \sa{i}\st{i} & a_i \ct{i} \\
            -\sa{i}\ct{i} & a_i\st{i} \\
            \ca{i} & d_i \\
            0 & 1
          \end{array}
        \right)
        ,
      \end{split}
    \end{multline}
where the joint parameters are given in \Cref{tab:DH-RViz}.    

\begin{table}[t]
  \centering
  \caption{Parameters in $A_i$ for the RViz coordinate system in myCobot}
  \begin{tabular}{cccccc}
    \hline
    Joint & $a_i$ & $\alpha_i$ & $d_i$ & $\theta_i$ & $\delta_i$\\
    \hline\hline
    1 & 0 & $\pi/2$ & $d_1$ & $\theta_1$ & $-\pi/2$\\
    2 & $-a_2$ & 0 & 0 & $\theta_2$ & 0\\
    3 & $-a_3$ & 0 & 0 & $\theta_3$ & $-\pi/2$\\
    4 & 0 & $\pi/2$ & $d_4$ & $\theta_4$ & $\pi/2$\\
    5 & 0 & $-\pi/2$ & $d_5$ & $\theta_5$ & 0\\
    6 & 0 & 0 & $d_6$ & $\theta_6$ & 0\\
    \hline
  \end{tabular}
  \label{tab:DH-RViz}
\end{table}

Let $A$ be the transformation matrix
from the end-effector's coordinate system $\Sigma_7$ to the global coordinate system $\Sigma_1$.
Then, $A$ is expressed as
\begin{equation}
  \label{eq:transformation-matrix}
  A=A_1A_2A_3A_4A_5A_6.
\end{equation}
\subsection{Description of the Orientation of the End-effector}
\label{sec:description-orientation}

Among the various methods to describe the orientation of the end-effector, this paper uses Roll-Pitch-Yaw angles \cite{sic-sci-vil-ori2008}.
Let $\bm{l},\bm{m},\bm{n}$ be the unit vectors of the $x,y,z$-axis of the end-effector, respectively, as
\begin{equation}
  \label{eq:orientation}
  \bm{l} = {}^t(l_1,l_2,l_3),\quad
  \bm{m} = {}^t(m_1,m_2,m_3),\quad
  \bm{n} = {}^t(n_1,n_2,n_3).
\end{equation}
Then, the orientation of the end-effector is expressed as
$\begin{pmatrix}
  \bm{l} & \bm{m} & \bm{n}
\end{pmatrix}
$.
To express the orientation of the end-effector with the variables of 3-DOF,
we use a series of the following rotations.
First, rotate around the $z$-axis by $\gamma$ (yaw angle), then rotate around the $y$-axis by $\beta$ (pitch angle). Finally, rotate around the $x$-axis by $\alpha$ (roll angle) to represent any orientation in 3D space.
Multiplying the rotation matrices of these rotations in order, we have a rotation matrix as
    \begin{multline}
      \label{eq:rpy}
      \begin{pmatrix}
        \bm{l} & \bm{m} & \bm{n}
      \end{pmatrix}
      =
      \\
      \begin{pmatrix}
        \cbe{}\cga{} & \cbe{}\sga{} & -\sbe{} \\
        \sa{}\sbe{}\cga{}-\ca{}\sga{} & \sa{}\sbe{}\sga{}+\ca{}\cga{} & \sa{}\cbe{} \\
        \ca{}\sbe{}\cga{}+\sa{}\sga{} & \ca{}\sbe{}\sga{}-\sa{}\cga{} & \ca{}\cbe{}
      \end{pmatrix}
      .
    \end{multline}
\subsection{Description of the Inverse Kinematic Problem of MyCobot}
\label{sec:description-inverse-kinematics}

Let the orientation $\bm{l},\bm{m},\bm{n}$ of myCobot be given as in \cref{eq:orientation} and
$A$ be the transformation matrix in \cref{eq:transformation-matrix}.
Let $\bm{p}={}^t(p_1,p_2,p_3)$ be the position of the end-effector w.r.t. $\Sigma_1$.
Since $\bm{l},\bm{m},\bm{n}$ are the unit vectors of the $x,y,z$-axis of $\Sigma_7$, respectively,
and $\bm{p}$ is the position of the end-effector with respect to $\Sigma_1$,
we have $A$ and $A^{-1}$ as
\begin{equation}
  \label{eq:inverse-kinematics}
    A =
    \begin{pmatrix}
      l_1 & m_1 & n_1 & p_1 \\
      l_2 & m_2 & n_2 & p_2 \\
      l_3 & m_3 & n_3 & p_3 \\
      0 & 0 & 0 & 1
    \end{pmatrix}
    ,
    \quad
    A^{-1} =
    \begin{pmatrix}
      l_1&l_2&l_3&-(l_1p_1+l_2p_2+l_3p_3) \\
      m_1&m_2&m_3&-(m_1p_1+m_2p_2+m_3p_3) \\
      n_1&n_2&n_3&-(n_1p_1+n_2p_2+n_3p_3) \\
      0&0&0&1
    \end{pmatrix}
    .
\end{equation}
which establishes the inverse kinematic equation of myCobot.
For given $\bm{l},\bm{m},\bm{n}$, and $\bm{p}$, the inverse kinematic problem is to find the joint angles $\theta_1,\dots,\theta_6$.
\section{Solving the Inverse Kinematic Problem of MyCobot}
\label{sec:solving-inverse-kinematics}

In this section, we propose a solution to myCobot's inverse kinematics problem.
First, we define the necessary notation.
\begin{itemize}
  \item Vectors are represented in bold, such as $\bm{P}$ or $\bm{w}$.
  \item A vector in the coordinate system $\Sigma_i$ is represented as ${}^i\bm{P}$. If it is clear from the context that the vector is in $\Sigma_1$, the superscript is unnecessary.
  \item A vector ${}^i\bm{P}$ is represented as ${}^j[{}^i\bm{P}]$ w.r.t. the coordinate system $\Sigma_j$.
  \item For a vector $\bm{P}$, the $i$th element is represented as $\bm{P}[i]$.
\end{itemize}
\subsection{Conventional Algebraic Method and Pieper's Approach}
\label{sec:conventional-algebraic-method}

To solve the inverse kinematics problem \eqref{eq:inverse-kinematics}, substitute the components of the end-effector's orientation and position into the right-hand side and compare the corresponding elements of the matrices on both sides.
Then, we obtain a system of 12 polynomial equations with respect to $\st{i}$ and $\ct{i}$ for $i=1,\dots,6$.
By appropriately selecting 6 of these equations and substituting $\st{i}$ and $\ct{i}$ with
$t_i=\tan\frac{\theta_i}{2}$, where $\st{i}=\frac{2t_i}{1+t_i^2}$ and
$\ct{i}=\frac{1-t_i^2}{1+t_i^2}$, we obtain six polynomial equations with six variables $t_i$. By eliminating one variable at a time, we derive a univariate polynomial equation from which each variable can be solved with backward substitution.
However, according to Pieper \cite{pie1968}, the degree of the resulting univariate equation exceeds $500000$, making it difficult to solve the inverse kinematics problem of a 6-DOF manipulator using algebraic methods in general.
Manocha and Canny \cite{man-can1994} have proposed a method to solve the inverse kinematics problem of a general 6-DOF manipulator.

Generally, a 6-DOF manipulator cannot independently determine the orientation and position of the end-effector.
Thus, it is difficult to solve the general inverse kinematics problem analytically.
\emph{However, if the manipulator includes a spherical joint with 3-dimensional rotation, the orientation and position of the end-effector can be determined independently.}
Then, the inverse kinematics problem can be separated into a 3-dimensional problem for orientation and a 3-dimensional problem for position.
In this case, the inverse kinematics problem will likely be solved analytically.

Pieper \cite{pie1968} has demonstrated that if the rotational axes of three consecutive rotational joints intersect at a single point, the motion of these three joints is kinematically equivalent to a spherical joint. He has proposed a solution to the inverse kinematics problem, focusing on this intersection point.
Unfortunately, in myCobot, the rotational axes of consecutive joints that intersect at a single point are Joint 1 and Joint 2, Joint 4 and Joint 5, and Joint 5 and Joint 6, meaning there is no intersection point for the rotational axes of three joints, which means that Pieper's method cannot be directly applied.
However, we have solved myCobot's inverse kinematics problem based on his approach.
Specifically, we express the intersection point of the rotational axes of two joints by the position and the orientation of the end-effector, and then derive the rotation angles of the joints using this intersection point.
\subsection{Deriving the Rotation Angle of the Joints}
\label{sec:deriving-rotation-angle}

Let $P$ be the intersection point of the rotational axes of Joints 4 and 5 as
\begin{equation}
  \label{eq:P}
  \bm{P} = {}^t(x,y,z).
\end{equation}
Note that $\bm{P}$ is at the origin of $\Sigma_5$.
We express $\st{i}$ and $\ct{i}$ in terms of $x,y,z$ for $i=1,\dots,6$.

\begin{remark}
  In solving the inverse kinematic problem, we use the transformation matrix $A$ and its inverse $A^{-1}$ in \eqref{eq:inverse-kinematics} and the 
orientation of the end-effector in \eqref{eq:orientation}.
  Note that the orientation of the end-effector is not represented using the Roll-Pitch-Yaw angles but rather by the components $l_i,m_i$, 
and $n_i$ for $i=1,2,3$.
  Furthermore, since $\bm{l},\bm{m}$, and $\bm{n}$ are the unit vectors in the direction of each axis of $\Sigma_7$, note that
  the following relationships hold.
  \begin{equation}
    \label{eq:lmn}
    \begin{split}
    &\quad l_2m_3-l_3m_2 =n_1,
    \quad l_3m_1-l_1m_3=n_2,
    \quad l_1m_2-l_2m_1=n_3, \\
    &\quad l_1^2+m_1^2+n_1^2 =1,
    \quad l_2^2+m_2^2+n_2^2=1,
    \quad l_3^2+m_3^2+n_3^2=1, \\
    &\quad l_1l_2+m_1m_2+n_1n_2=0,
    \quad l_2l_3+m_2m_3+n_2n_3=0, \\
    &\quad l_3l_1+m_3m_1+n_3n_1=0.
    \end{split}
  \end{equation}
\end{remark}

Now, for each $i=1,\dots,6$, determine $\st{i}$ and $\ct{i}$ using the coordinates $x,y,z$ of the intersection point $\bm{P}$.
\subsubsection{Deriving $\st{6}$ and $\ct{6}$}
\label{sec:deriving-stct6}

The position vector ${}^7\bm{P} = \overrightarrow{\origin{7}P}$ is expressed in the following two ways.
The intersection point $\bm{P}$ is located at the origin of $\Sigma_5$, so it can be expressed as
\begin{equation}
  \label{eq:7P-1}
    {}^7\bm{P} = A_6^{-1}A_5^{-1}\;{}^t(0,0,0,1)
     = {}^t({d_5\st{6}},{d_5\ct{6}},{-d_6},1).
\end{equation}
On the other hand, since $\bm{P}={}^t(x,y,z,1)$, it can be expressed as
\begin{equation}
  \label{eq:7P-2}
  \begin{split}
    {}^7\bm{P} =
    A^{-1}
    \Vector{x}{y}{z}{1}
    = \Vector{l_1(-p_1+x)+l_2(-p_2+y)+l_3(-p_3+z)}{m_1(-p_1+x)+m_2(-p_2+y)+m_3(-p_3+z)}{n_1(-p_1+x)+n_2(-p_2+y)+n_3(-p_3+z)}{1}.
  \end{split}
\end{equation}
Since each element in the rightmost-hand-side of \cref{eq:7P-1,eq:7P-2} is equal, we have
\begin{equation}
  \label{eq:stct6}
  \begin{split}
  \st{6} &= \frac{l_1(-p_1+x)+l_2(-p_2+y)+l_3(-p_3+z)}{d_5},\\
  \ct{6} &= \frac{m_1(-p_1+x)+m_2(-p_2+y)+m_3(-p_3+z)}{d_5}.
  \end{split}
\end{equation}
\subsubsection{Deriving $\st{5}$ and $\ct{5}$}
\label{sec:deriving-stct5}

Let $\bm{w}$ be the unit vector in the direction of the joint axis of Joint 4.
Then, $\bm{w}_4$ and ${}^7[\bm{w}_4]$ are expressed as
\begin{equation}
  \label{eq:w4-1}
  \begin{split}
    \bm{w}_4
    &= A_1A_2A_3\;
    {}^t(0,0,1,0)
    = {}^t({\st{1}},{-\ct{1}},0,0),\\
    {}^7\left[\bm{w}_4\right]
    & =A_6^{-1}A_5^{-1}A_4^{-1}\,{}^t(0,0,1,0)
    \\
    &
    = {}^t({\ct{5}\ct{6}},{-\ct{5}\st{6}},{-\st{5}},0).
  \end{split}
\end{equation}
On the other hand, we also have
\begin{equation}
  \label{eq:w4-2}
  \bm{w}_4=A\cdot{}^7\left[\bm{w}_4\right]
  =\Vector{-n_1\st{5}+\ct{5}(l_1\ct{6}-m_1\st{6})}{-n_2\st{5}+\ct{5}(l_2\ct{6}-m_2\st{6})}
  {-n_3\st{5}+\ct{5}(l_3\ct{6}-m_3\st{6})}{0}.
\end{equation}
Since each element of $\bm{w}_4$ in \cref{eq:w4-1,eq:w4-2} is equal.
By comparing the third element, we have
\begin{equation}
  \label{eq:w4-3}
  -n_3\st{5}+\ct{5}(l_3\ct{6}-m_3\st{6})=0.
\end{equation}
By substituting $\st{6}$ and $\ct{6}$ from \cref{eq:stct6} into \cref{eq:w4-3} and simplifying using \cref{eq:lmn}, we have
\begin{equation}
  \label{eq:stct5-ren}
  \begin{split}
    d_5n_3\st{5}+\ct{5}(n_2(p_1-x)-n_1(p_2-y)) &=0, \\
    (\st{5})^2+(\ct{5})^2 &=1.
  \end{split}
\end{equation}
In the case $n_3\neq0$ or $n_2(p_1-x)-n_1(p_2-y)\neq 0$, by \cref{eq:stct5-ren},
$\st{5}$ and $\ct{5}$ are expressed as
\begin{equation}
  \label{eq:stct5}
  \begin{split}
    \st{5} &= \pm
    \frac{n_2(p_1-x)-n_1(p_2-y)}{\sqrt{d_5^2n_3^2+(n_2(p_1-x)-n_1(p_2-y))^2}}
    ,\\
    \ct{5} &= \mp
    \frac{d_5n_3}{\sqrt{d_5^2n_3^2+(n_2(p_1-x)-n_1(p_2-y))^2}}.
  \end{split}
\end{equation}
In the case where $n_3=0$ and $n_2(p_1-x)-n_1(p_2-y) = 0$, it is described in
\Cref{sec:derivation-special}.
\subsubsection{Deriving $\st{1}$ and $\ct{1}$}
\label{sec:deriving-stct1}

For $\st{1}$ and $\ct{1}$, a comparison of the first and the second components of $\bm{w}_4$ in \cref{eq:w4-1,eq:w4-2} shows that
\begin{equation}
  \label{eq:stct1-ren}
  \begin{split}
    \st{1} &= -n_1\st{5}+\ct{5}(l_1\ct{6}-m_1\st{6}), \\
    \ct{1} &= n_2\st{5}-\ct{5}(l_2\ct{6}-m_2\st{6}).
  \end{split}
\end{equation}
By substituting $\st{6},\ct{6},\st{5},\ct{5}$ from \cref{eq:stct6,eq:stct5} and using
\cref{eq:lmn}, $\st{1}$ and $\ct{1}$ are expressed as
\begin{equation}
  \label{eq:stct1}
  \begin{split}
  \st{1} &= \pm
  \frac{-n_1n_2(p_1-x)+(1-n_2^2)(p_2-y)-n_2n_3(p_3-z)}{\sqrt{d_5^2n_3^2+(n_2(p_1-x)-n_1(p_2-y))^2}}
  ,\\
  \ct{1} &= \pm
  \frac{(1-n_1^2)(p_1-x)-n_1n_2(p_2-y)-n_1n_3(p_3-z)}{\sqrt{d_5^2n_3^2+(n_2(p_1-x)-n_1(p_2-y))^2}}.
  \end{split}
\end{equation}
\subsubsection{Deriving $\st{3}$ and $\ct{3}$}
\label{sec:deriving-stct3}

Let $\bm{P}_3=\overrightarrow{\origin{1}P}$.
Since $\bm{P}$ is the origin of $\Sigma_5$, $\bm{P}_3$ satisfies
$\bm{P}_3=\overrightarrow{\origin{1}\origin{5}}$ as
    \begin{equation}
      \label{eq:P3-1}
      \begin{split}
      &\quad \bm{P}_3 =A_1A_2A_3A_4\; {}^t(0,0,0,1) \\
      &=
      \begin{pmatrix}
      d_4\st{1}-a_2\ct{1}\st{2}-a_3\ct{1}(\st{2}\ct{3}+\ct{2}\st{3})\\
      -d_4\ct{1}-a_2\st{1}\st{2}-a_3\st{1}(\st{2}\ct{3}+\ct{2}\st{3})\\
      d_1+a_2\ct{2}+a_3(\ct{2}\ct{3}-\st{2}\st{3})\\
      1
      \end{pmatrix}
      .
      \end{split}
    \end{equation}
On the other hand, since the coordinate of $\bm{P}$ is ${}^t(x,y,z)$, $\bm{P}_3$ is expressed as
\begin{equation}
  \label{eq:P3-2}
  \bm{P}_3 = {}^t(x,y,z,1).
\end{equation}
Let $Q=\bm{P}_3[1]^2+\bm{P}_3[2]^2+(\bm{P}_3[3]-d_1)^2$.
By calculating $Q$ using the components in \cref{eq:P3-1,eq:P3-2}, we have
\begin{equation}
  \label{eq:Q}
  a_2^2+a_3^2+d_4^2+2a_2a_3\ct{3} = x^2+y^2+(z-d_1)^2.
\end{equation}
Thus, $\st{3}$ and $\ct{3}$ are expressed as
\begin{equation}
  \label{eq:stct3}
  \begin{split}
    \st{3}= \pm \sqrt{1-(\ct{3})^2},\quad
    \ct{3}= \frac{x^2+y^2+(z-d_1)^2-a_2^2-a_3^2-d_4^2}{2a_2a_3}.
  \end{split}
\end{equation}
\subsubsection{Deriving $\st{2}$ and $\ct{2}$}
\label{sec:deriving-stct2}

By comparing the third component of $\bm{P}_3$ in \cref{eq:P3-1,eq:P3-2},
we have
\begin{equation}
  \label{eq:stct2}
  \begin{split}
    d_1-a_2\ct{2}-a_3(\ct{2}\ct{3}-\st{2}\st{3}) &= z, \\
    (\st{2})^2+(\ct{2})^2 &=1.
  \end{split}
\end{equation}
Thus, $\st{2}$ and $\ct{2}$ can be obtained by substituting the expressions for $\st{3}$ and $\ct{3}$ from \cref{eq:stct3} into \cref{eq:stct2}.
\subsubsection{Deriving $\st{4}$ and $\ct{4}$}
\label{sec:deriving-stct4}

Let $\bm{w}_5$ be the unit vector in the direction of the joint axis of Joint 5.
Then we have ${}^5[\bm{w}_5]={}^t(0,0,1,0)$.
Thus, $\bm{w}_5$ and ${}^7[\bm{w}_5]$ are expressed as
\begin{equation}
  \label{eq:w5-1}
  \begin{split}
    \bm{w}_5 &= A_1A_2A_3A_4\Vector{0}{0}{1}{0}
    = \Vector
    {-\ct{1}\sin(\theta_2+\theta_3+\theta_4)}
    {-\st{1}\sin(\theta_2+\theta_3+\theta_4)}
    {\cos(\theta_2+\theta_3+\theta_4)}
    {0},\\
    {}^7\left[\bm{w}_5\right] &=
    A_6^{-1}A_5^{-1}\; {}^t(0,0,1,0)
    = {}^t({-\st{6}},{-\ct{6}},{0},{0}).
  \end{split}
\end{equation}
On the other hand, we also have
\begin{equation}
    \label{eq:w5-2}
    \bm{w}_5 =
    A\cdot{}^7\left[\bm{w}_5\right]
    = \Vector{-m_1\ct{6}-l_1\st{6}}{-m_2\ct{6}-l_2\st{6}}{-m_3\ct{6}-l_3\st{6}}{0}.
\end{equation}
Since each element in the rightmost-hand-side of $\bm{w}_5$ in \cref{eq:w5-1,eq:w5-2} is equal,
by comparing the third element, we have
\begin{equation}
  \label{eq:stct4}
  \begin{split}
    \cos(\theta_2+\theta_3+\theta_4)
    &=(\ct{2}\ct{3}-\st{2}\st{3})\ct{4}
    \\
    &\quad -(\st{2}\ct{3}+\ct{2}\st{3})\st{4}
    \\
    &=-m_3\ct{6}-l_3\st{6},
    \\
    (\ct{4})^2+(\st{4})^2 &= 1.
    \end{split}
\end{equation}
Furthermore, $\st{4}$ and $\ct{4}$ can be obtained by substituting the expressions for
$\st{i}$ and $\ct{i}$ for $i=1$ (in \cref{eq:stct1}), $2$ (in \cref{eq:stct2}), $3$ (in \cref{eq:stct3}), and $6$ (in \cref{eq:stct6}) into \cref{eq:stct4}.

\subsection{Deriving the Intersection Point}
\label{sec:deriving-P}

We derive $x,y$, and $z$ of the intersection point $\bm{P}$ in \cref{eq:P}.
First, by comparing the third component of the vector ${}^7\bm{P}$
from \cref{eq:7P-1,eq:7P-2}, we have
\begin{equation}
  \label{eq:xyz-1}
  n_1(p_1-x)+n_2(p_1-y)+n_3(p_3-z)=d_6.
\end{equation}

Next, comparing the third component of the vector
${}^7\bm{P}$ from \cref{eq:7P-1,eq:7P-2}, 
we wee that 
    \begin{equation}
      \label{eq:7P-norm}
      \begin{split}
      \|{}^7\bm{P}\|^2 &=d_5^2+d_6^2,\\
      \|{}^7\bm{P}\|^2
      &=\left(l_1(-p_1+x)+l_2(-p_2+y)+l_3(-p_3+z)\right)^2\\
      &\quad +\left(m_1(-p_1+x)+m_2(-p_2+y)+m_3(-p_3+z)\right)^2\\
      &\quad +\left(n_1(-p_1+x)+n_2(-p_2+y)+n_3(-p_3+z)\right)^2\\
      &= (l_1^2+m_1^2+n_1^2)(p_1-x)^2+(l_2^2+m_2^2+n_2^2)(p_2-y)^2\\
      &\quad +(l_3^2+m_3^2+n_3^2)(p_3-z)^2\\
      &\quad
      +2(l_1l_2+m_1m_2+n_1n_2)(p_1-x)(p_2-y)\\
      &\quad +2(l_2l_3+m_2m_3+n_2n_3)(p_2-y)(p_3-z)\\
      &\quad+2(l_3l_1+m_3m_1+n_3n_1)(p_3-z)(p_1-x)\\
      &=(p_1-x)^2+(p_2-y)^2+(p_3-z)^2.
      \end{split}
    \end{equation}
Thus, we have
\begin{equation}
  \label{eq:xyz-2}
  (p_1-x)^2+(p_2-y)^2+(p_3-z)^2 = d_5^2+d_6^2.
\end{equation}

Thirdly, let ${}^5\bm{\origin{1}}=\overrightarrow{P\origin{1}}$.
Considering it as a position vector, using the transformation matrices $
A_1^{-1}, A_2^{-1}, A_3^{-1}, A_4^{-1}$, it can be expressed as
\begin{equation}
  \label{eq:5O-1}
  \begin{split}
		{}^5\bm{\origin{1}} &= A_4^{-1}A_3^{-1}A_2^{-1}A_1^{-1}
    \cdot {}^t(0,0,0,1)\\
		&=
		\begin{pmatrix}
			-d_4 \\
			-a_3\st{4}-a_2\sin(\theta_3+\theta_4)-d_1\sin(\theta_2+\theta_3+\theta_4) \\
		-a_3\ct{4}-a_2\cos(\theta_3+\theta_4)-d_1\cos(\theta_2+\theta_3+\theta_4) \\
			1
		\end{pmatrix}
    .
	\end{split}
\end{equation}
Additionally, the vector $\overrightarrow{P\origin{1}}$ and the vector $\overrightarrow{\origin{1}P}$ are vectors in opposite directions. Therefore, in the coordinate system $\Sigma_1$, the direction vector
${}^1[\overrightarrow{P\origin{1}}]={}^1[{}^5\bm{\origin{1}}] = (-x, -y, -z, 0)^T$ holds. Consequently, ${}^5\bm{\origin{1}}$ can also be expressed as
\begin{equation}
  \label{eq:5O-2}
  \begin{split}
    {}^5\bm{\origin{1}} &= A_4^{-1}A_3^{-1}A_2^{-1}A_1^{-1}\cdot{}^1[{}^5\bm{\origin{1}}]
    ={}^t(\origin{1,1},\origin{1,2},\origin{1,3},0),\\
    \origin{1,1} &= y\ct{1}-x\st{1}, \\
    \origin{1,2} &= \frac{1}{2}(-x(\ct{1-2-3-4}+\ct{1+2+3+4})\\
    &\qquad -y(\st{1-2-3-4}+\st{1+2+3+4})-2z\st{2+3+4}),\\
    \origin{1,3} &= \frac{1}{2}(-x(\st{1-2-3-4}-\st{1+2+3+4})\\
    &\qquad +y(\ct{1-2-3-4}-\ct{1+2+3+4})-2z\ct{2+3+4}). \\
  \end{split}
\end{equation}
where
    $\theta_{1-2-3-4} = \theta_1-\theta_2-\theta_3-\theta_4$,
    $\theta_{1+2+3+4} = \theta_1+\theta_2+\theta_3+\theta_4$,
    $\theta_{2+3+4} = \theta_2+\theta_3+\theta_4$.

The first, second, and third components of ${}^5\bm{\origin{1}}$ in \cref{eq:5O-1,eq:5O-2}
are equal, respectively. The comparison of the first component shows that
\begin{equation}
  \label{eq:yct1-xst1}
  y\ct{1}-x\st{1} = -d_4.
\end{equation}
By substituting \cref{eq:stct1} into \cref{eq:yct1-xst1} and simplifying, the following holds.
\begin{equation*}
  \small
  \begin{split}
  \quad\pm d_4 &=
  y
  \frac
  {(1-n1^2)(p_1-x)-n_1n_2(p_2-y)-n_1n_3(p_3-z)}
  {\sqrt{d_5^2n_3^2+(n_2(p_1-x)-n_1(p_2-y))^2}}
  \\
  &\quad -x
  \frac
  {-n_1n_2(p_1-x)+(1-n_2^2)(p_2-y)-n_2n_3(p_3-z)}
  {\sqrt{d_5^2n_3^2+(n_2(p_1-x)-n_1(p_2-y))^2}},
\end{split}
\end{equation*}
thus we have
\begin{equation}
  \label{eq:xyz-3}
  \begin{split}
    & d_4^2(d_5^2n_3^2+(n_2(p_1-x)-n_1(p_2-y))^2) \\
    =\; &((n_1n_2x+(1-n_1^2)y)(p_1-x)
    -(n_1n_2y+(1-n_2^2)x)(p_2-y)\\
    &\quad -n_3(n_1y-n_2x)(p_3-z))^2.
  \end{split}
\end{equation}
Therefore, the following system of parametric algebraic equations for $x,y$, and $z$ is derived from equations \cref{eq:xyz-1,eq:xyz-2,eq:xyz-3}, as
\begin{equation}
  \label{eq:CGS}
  \begin{split}
    & n_1(p_1-x)+n_2(p_2-y)+n_3(p_3-z)=d_6,\\
    & (p_1-x)^2+(p_2-y)^2+(p_3-z)^2=d_5^2+d_6^2,\\
    & ((n_1n_2x+(1-n_1^2)y)(p_1-x)\\
    &\quad -(n_1n_2y+(1-n_2^2)x)(p_2-y)-n_3(n_1y-n_2x)(p_3-z) )^2 \\
    &\qquad\qquad =d_4^2(d_5^2n_3^2+(n_2(p_1-x)-n_1(p_2-y))^2),
  \end{split}
\end{equation}
where parameters are $n_1,n_2$, and $n_3$ (direction of the end-effector's $z$-axis),
$p_1,p_2$, and $p_3$ (position of the end-effector).
Note that $d_4,d_5$, and $d_6$ are the lengths of the links in the DH parameter (see \cref{tab:DH-RViz}).
When actually solving \eqref{eq:CGS},
the lengths of the links are assigned to $d_4,d_5$, and $d_6$ in terms of rational numbers.

When the orientation and position of the end-effector are given,
by solving \cref{eq:CGS}, the coordinates $x,y,z$ of the intersection point $\bm{P}$ can be determined.
Then, $\st{i},\ct{i}$ $(i=1,\dots,6)$ can be obtained using the method from above, allowing $\theta_i$ to be determined.

To solve \cref{eq:CGS}, one can compute the Gr\"obner basis after substituting the orientation and position of the end-effector.
However, the computation of the Gr\"obner basis each time after the substitution is time-consuming. Therefore, by following the approach of our previous research \cite{ota-ter-mik2021,yos-ter-mik2023}, computing the CGS of the ideal generated by the polynomials in \cref{eq:CGS}, the corresponding Gr\"obner basis can be quickly obtained once the orientation and position of the end-effector are given, significantly reducing the time required.

\begin{remark}
  \label{rem:check-results}
  In \cref{eq:CGS}, some equations are squared on both sides to eliminate complex signs and radicals, making it easier to compute the CGS.
  Then, the solution $x,y,z$ of the intersection point $\bm{P}$ and the solutions to the inverse kinematics problem may include values that differ from the original solutions.
  Thus, after solving the inverse kinematics problem, it is necessary to solve the forward kinematic problem to verify the results.
\end{remark}
\subsection{Deriving and Solving the Inverse Kinematics Problem for Special Orientations}
\label{sec:derivation-special}

The above method cannot be used when the end-effector's orientation is horizontal to the ground, i.e., when $n_3=0$.
In the case $n_3=0$, problems arise in \cref{eq:stct5-ren,eq:CGS}.
We discuss these issues and their solutions.
\subsubsection{In the Case $n_3=0$ and $n_2(p_1-x)-n_1(p_2-y) = 0$}
\label{sec:problem-stct5}

When certain $x,y$ satisfy $n_3=0$ and $n_2(p_1-x)-n_1(p_2-y) = 0$,
the equation $d_5n_3\st{5}+\ct{5}(n_2(p_1-x)-n_1(p_2-y))$ in \cref{eq:stct5-ren} becomes identically zero.
Consequently, $\st{5}$ and $\ct{5}$ cannot be determined, and $\st{1}$, $\ct{1}$ derived from $\st{5}$, $\ct{5}$, as well as \cref{eq:xyz-3}, cannot be obtained.
Therefore, a new method for determining $\st{1}$, $\ct{1}$, $\st{5}$, $\ct{5}$, and the coordinates $x,y,z$ of the intersection point $P$ is established.

We first calculate $x,y$, and $z$. From the assumptions of this case, we have
\begin{equation}
  \label{eq:xyz-4}
  n_2(p_1-x)-n_1(p_2-y) = 0.
\end{equation}
Furthermore, \cref{eq:xyz-1,eq:xyz-2} hold regardless of the current assumptions.
Therefore, the values of $x,y,z$ that vanish $d_5n_3\st{5}+\ct{5}(n_2(p_1-x)-n_1(p_2-y))$ in \cref{eq:stct5-ren} can be determined from the following system of parametric equations:
\begin{equation}
  \label{eq:CGS_n3_0_1}
  \begin{split}
    & n_1(p_1-x)+n_2(p_2-y)=d_6,\\
    & (p_1-x)^2+(p_2-y)^2+(p_3-z)^2=d_5^2+d_6^2,\\
    & n_2(p_1-x)-n_1(p_2-y) = 0,
  \end{split}
\end{equation}
where parameters are $n_1$ and $n_2$ (direction of the end-effector's $z$-axis), $p_1,p_2$, and $p_3$ (position of the end-effector).
Note that $d_5,d_6$ are the lengths of the links in the DH parameter and,
in actual calculations, real numbers are substituted.


Next, for $\st{1}$ and $\ct{1}$, by comparing the first component of $\bm{P}_3$ from
\cref{eq:P3-1,eq:P3-2}, we have
\begin{equation}
  \label{eq:stct1_1}
  \begin{split}
    & d_4\st{1}-a_2\ct{1}\st{2}
    -a_3\ct{1}(\st{2}\ct{3}+\ct{2}\st{3}) = x,\\
    & (\st{1})^2+(\ct{1})^2 = 1.
  \end{split}
\end{equation}

Once $x,y$, and $z$ are determined from \cref{eq:CGS_n3_0_1}, $\st{2},\ct{2},\st{3}$, and $\ct{3}$ can be obtained.
By substituting the obtained results into \cref{eq:stct1_1}, $\st{1}$ and $\ct{1}$ can be determined.

For $\st{5}$ and $\ct{5}$, by comparing the first component of $\bm{w}_4$ from
\cref{eq:w4-1,eq:w4-2}, we have
\begin{equation}
  \label{eq:stct5_1}
  \begin{split}
    -n_1\st{5}+\ct{5}(l_1\ct{6}-m_1\st{6}) &= \st{1},\\
    -n_2\st{5}+\ct{5}(l_2\ct{6}-m_2\st{6}) &= -\ct{1}.
  \end{split}
\end{equation}
$\st{6}$ and $\ct{6}$ can be obtained in the same way as in \Cref{sec:deriving-stct6}.
$\st{1}$ and $\ct{1}$ can be determined from \cref{eq:stct1_1}.
By substituting these into \cref{eq:stct5_1}, $\st{5}$ and $\ct{5}$ can be obtained.
\subsubsection{In the Case $n_3=0$ and $n_2(p_1-x)-n_1(p_2-y)\ne 0$}
\label{sec:problem-CGS}

Assume that $n_3=0$ and $n_2(p_1-x)-n_1(p_2-y)\ne 0$.
By putting $n_3=0$ into \cref{eq:xyz-1}, we have
$n_1(p_1-x)+n_2(p_2-y) = d_6$,
which derives
  $n_1x+n_2y = n_1p_1+n_2p_2-d_6$.
Next, by substituting $n_3=0$ into \cref{eq:xyz-3}, we have
\begin{multline*}
  (n_2(p_1-x)-n_1(p_2-y))^2(n_1x+n_2y)^2
  = d_4^2(n_2(p_1-x)-n_1(p_2-y))^2,
\end{multline*}
which derives
    $(n_1x+n_2y)^2 = d_4^2$, or
    $n_1x+n_2 = \pm d_4$.
We see that, for \cref{eq:CGS} to have a solution,
$n_1p_1+n_2p_2-d_6=\pm d_4$ must hold.
Additionally, even if $n_1p_1+n_2p_2-d_6=\pm d_4$ is satisfied,
there are only two equations for three variables, so the solution may not be unique.
Thus, it is necessary to find one more equation for $x,y$, and $z$.

By substituting $n_3=0$ into \cref{eq:stct1}, $\st{1}$ and $\ct{1}$ can be expressed as
\begin{equation}
  \label{eq:stct1_2}
  \st{1} = \pm n_1, \quad \ct{1} = \mp n_2.
\end{equation}
Furthermore, by substituting \cref{eq:stct1_2}
into the components of $\bm{P}_3$ in
\cref{eq:P3-1}, and comparing each component with those in \cref{eq:P3-2}, we have
\begin{equation}
  \label{eq:stct23-1}
  \begin{split}
    \pm d_4n_1\pm a_2n_2\st{2}\pm a_3n_2(\st{2}\ct{3}+\ct{2}\st{3}) &= x,\\
    \pm d_4n_2\mp a_2n_1\st{2}\mp a_3n_1(\st{2}\ct{3}+\ct{2}\st{3}) &= y,\\
    d_1+a_2\ct{2}+a_3(\ct{2}\ct{3}-\st{2}\st{3}) &= z,\\
    (\st{2})^2+(\ct{2})^2 = 1,\quad
    (\st{3})^2+(\ct{3})^2 &= 1.
  \end{split}
\end{equation}
In \eqref{eq:stct23-1}, solving equations 1, 3, 4, and 5 for $\st{2},\ct{2}$,
$\st{3}$, and $\ct{3}$ yields two sets of solutions. In this case, $\ct{3}$ is expressed as
\[
\ct{3} = \frac{d_4^2 n_1^2 - a_2^2 n_2^2 - a_3^2 n_2^2 + d_1^2 n_2^2 \mp 2 d_4 n_1 x + x^2 - 2 d_1 n_2^2 z + n_2^2 z^2}{2a_2a_3n_2^2},
\]
for both sets.
By substituting the above $\ct{3}$ into $\ct{3}$ in \cref{eq:Q}, the following holds.
In the case $n_2 \neq 0$, we have
\[
  \frac{d_4^2\mp2d_4n_1x+n_1^2x^2-n_2^2y^2}{n_2^2} = 0,
\]
which derives
\begin{equation}
  \label{eq:xyz-5}
  (d_4^2+n_1^2x^2-n_2^2y^2)^2-4d_4^2n_1^2x^2 = 0.
\end{equation}
In the case $n_2=0$, performing the same calculations as above for equations 1, 2, 4, and 5 in \eqref{eq:stct23-1}, we have
\begin{equation}
  \label{eq:xyz-6}
  (d_4^2-n_1^2x^2+n_2^2y^2)^2-4d_4^2n_2^2y^2 = 0.
\end{equation}
Thus, $x,y,z$ can be obtained by solving \cref{eq:CGS} along with either \cref{eq:xyz-5}
or \cref{eq:xyz-6}.

\subsection{An algorithm}
\label{sec:algorithm}

Based on the above results, the algorithm for solving the inverse kinematics problem of myCobot is shown in \Cref{alg:inverse} (Note that \Cref{alg:inverse} reflects the flow of calculations from
\cref{sec:deriving-rotation-angle} to \ref{sec:problem-CGS}).
In \Cref{alg:inverse}, the order in which each $\theta_i$ is determined is partially unordered.
However, note that we must calculate $\theta_3,\theta_2$, and $\theta_4$ in order in
\Cref{sec:deriving-rotation-angle},
and $\theta_3,\theta_2$, and $\theta_1$ in order in \Cref{sec:derivation-special}.
Furthermore, if multiple solutions are obtained for each $\theta_i$, it is necessary to determine the set $(\theta_1,\dots,\theta_6)$ for each case.

\begin{algorithm}[t]
  \caption{Solving the inverse kinematic problem in myCobot}
  \label{alg:inverse}
  \begin{algorithmic}[1]
    \Input{The end-effector's position $\bm{p}=(p_1,p_2,p_3)$ and orientation ($\alpha,\beta,\gamma$);}
    \Output{A set $K'$ of tuples of joint angles $(\theta_1,\dots,\theta_6)$;}
    \State{Calculate CGS of \cref{eq:CGS,eq:CGS_n3_0_1,};}
    \State{Calculate $l_i,m_i,n_i$ ($i=1,2,3$) by substituting $\alpha,\beta,\gamma$ into \cref{eq:rpy};}
    \label{alg:inverse:substitution}
    \State{$K\gets\varnothing$;\quad $K'\gets\varnothing$;}
    \If{$n_3=0$}
      \If{$n_2(p_1-x)-n_1(p_2-y)=0$}
        \State{Choose an appropriate segment from the CGS of \cref{eq:CGS_n3_0_1};}
      \Else
        \State{Choose an appropriate segment from the CGS of \cref{eq:CGS} along with \cref{eq:xyz-6} (for $n_2=0$) or \cref{eq:xyz-5} (for $n_2\ne 0$);}
      \EndIf
      \State{Substitute $l_i,m_i,n_i,p_i$ ($i=1,2,3)$ in the CGS with the calculated values;}
      \State{Calculate the intersection point $(x,y,z)$ from the selected Gr\"obner basis;}
      \State{Calculate $\theta_6,\theta_3,\theta_2,\theta_1,\theta_4,\theta_5$ sequentially from
      \eqref{eq:stct6}, \eqref{eq:stct3}, \eqref{eq:stct2}, \eqref{eq:stct1_1}, \eqref{eq:stct4} and \eqref{eq:stct5_1}, respectively;}
    \Else
      \State{Choose an appropriate segment from the CGS of \cref{eq:CGS};}
      \State{Substitute $l_i,m_i,n_i,p_i$ ($i=1,2,3)$ in the CGS with the calculated values;}
      \State{Calculate the intersection point $(x,y,z)$ from the selected Gr\"obner basis;}
      \State{Calculate $\theta_6,\theta_5,\theta_1,\theta_3,\theta_2,\theta_4$ sequentially from
      \eqref{eq:stct6}, \eqref{eq:stct5}, \eqref{eq:stct1}, \eqref{eq:stct3}, \eqref{eq:stct2} and  \eqref{eq:stct4}, respectively;}
    \EndIf
    \State{$K\gets K\cup\{(\theta_1,\theta_2,\theta_3,\theta_4,\theta_5,\theta_6)\}$;}
    \ForEach{$(\theta_1,\theta_2,\theta_3,\theta_4,\theta_5,\theta_6)\in K$}
      \State{Substitute $(\theta_1, \theta_2, \theta_3, \theta_4, \theta_5, \theta_6)$ into the transformation matrix $A$ of \cref{eq:transformation-matrix};}
      \If{$A = (\bm{l},\bm{m},\bm{n},\bm{p})$} \label{alg:inverse:check}
        \State{$K'\gets K'\cup\{(\theta_1,\theta_2,\theta_3,\theta_4,\theta_5,\theta_6)\}$;}
      \EndIf
    \EndFor
    \State{\Return $K'$.}
  \end{algorithmic}
\end{algorithm}
\section{Implementation and Experiments}
\label{sec:experiments}

In this section, we present an optimization of \Cref{alg:inverse}, an implementation and the experimental results of the proposed algorithm.
\subsection{Efficient Calculation of $\st{i}$ and $\ct{i}$ with Gr\"obner Basis 
Computation}
\label{sec:optimization}

\Cref{alg:inverse} performs the calculation of $\ct{i}$ and $\st{i}$ $(i=1,\dots,6)$,
according to the calculation described in \Cref{sec:solving-inverse-kinematics}.
For $i=1,\dots,6$, let $c_i=\cos\theta_i$ and $s_i=\sin\theta_i$.
In the generic case, the equations with respect to $s_i$ and $c_i$ are derived as a system of 12 polynomial equations
  $f_1=\cdots=f_{12}=0$,
derived from \cref{eq:stct6,eq:stct5-ren,eq:stct1-ren,eq:Q,eq:stct2}, and \eqref{eq:stct4}.
Then, for solving $f_1=\cdots=f_{12}=0$,
we calculate the Gr\"obner basis $G$ of the ideal $\langle f_1,\dots,f_{12}\rangle$
in
$\mathbb{R}(l_1,l_2,l_3$, $m_1,m_2,m_3,n_1,n_2,n_3,p_1,p_2,p_3,x,y,z)$ $[c_1,s_1,c_2,s_2,c_3,s_3,c_4,s_4,c_5,s_5,$ $c_6,s_6]$
with the lexicographic order $s_4>c_4>s_2>c_2>s_3>c_3>s_1>c_1>s_5>c_5>s_6>c_6$.

In the following experimental results,
the use of $G$ significantly reduced the computing time,
and it also improved the computation accuracy (see the experimental results in
\Cref{sec:accuracy-inverse}).

Note that, unless otherwise noted, all the following calculation results are obtained
using \Cref{alg:inverse} together with the pre-computed Gr\"obner basis $G$.
\subsection{Implementation}
\label{sec:implementation}

We have implemented \Cref{alg:inverse} on the computer algebra system Risa/Asir~\cite{nor-tak1992}.
The CGS calculations were performed using the implementation on Risa/Asir by Nabeshima \cite{nab2018}, which implements the algorithm by Kapur et al. \cite{kap-sun-wan2013}.
The functionality for computing numerical roots of univariate polynomial equations utilizes the capabilities of the computer algebra system PARI/GP 2.13.1
\cite{PARI2.13.1}, which is invoked as a built-in function from Risa/Asir.

The computing environment is as follows:
Intel Xeon Silver 4210 3.2 GHz,
RAM 256 GB,
Linux Kernel 5.4.0,
Risa/Asir Version 20230315.

\begin{remark}
  As we have mentioned in \Cref{rem:check-results}, after solving the inverse kinematics problem, it is necessary to solve the forward kinematic problem to verify the results.
  While this procedure is included in line \ref{alg:inverse:check} of \Cref{alg:inverse},
  in our implementation, for efficiency, only the position of the end-effector $\bm{p}$ is inspected.
\end{remark}
\subsection{Calculating the CGS}
\label{sec:CGS}

In this section, we show the calculation results of the CGS for the parametric systems of
equations obtained in \Cref{sec:solving-inverse-kinematics}.
We have calculated the CGS for the following cases of parametric systems of equations
after substituting the DH parameters in \Cref{tab:DH-RViz}:
Case 1: \cref{eq:CGS};
Case 2: substituting $n_1=0,n_2=0,n_3=\pm1$ in \cref{eq:CGS};
Case 3: \cref{eq:CGS_n3_0_1};
Case 4: \cref{eq:CGS} along with \cref{eq:xyz-5};
and Case 5: \cref{eq:CGS} along with \cref{eq:xyz-6}.

\Cref{tab:cgs} shows the computing time (in seconds) and the number of segments%
\footnote{Due to the lack of space, we refer to references (e.g., \cite{suz-sat2006}) for segments and other key concepts of the CGS.}
of the CGS obtained.
Although the calculation for \cref{eq:CGS} continued for about two months, a complete
CGS was not obtained.
However, the first segment and its corresponding Gr\"ober basis was obtained in approximately 2.59 seconds, suggesting that the calculation of the other segments is time-consuming.
\begin{table}[t]
  \centering
  \caption{Computing time of CGS}
  \label{tab:cgs}
  \begin{tabular}{c|cc}
  \hline
  Case & Time [s] & \# Segments\\
  \hline
  1 & $>2$ Months & $>1$ \\
  1 (the 1st segment) & 2.58713 & 1 \\
  2 & 0.021623 & 9 \\
  3 & 0.001202 & 2 \\
  4 & 11.2487 & 11 \\
  5 & 10.0062 & 11 \\
  \hline
  \end{tabular}
\end{table}

In our implementation of \Cref{alg:inverse}, since only one segment of the CGS has been calculated,
we assume that the position and orientation of the end effector in the input belong to the only
calculated segment, and substitute them into the coefficients of the Gr\"obner basis to calculate
the intersection point.
\subsection{Accuracy of the  Inverse  Kinematic                                 Solution}
\label{sec:accuracy-inverse}

We first conducted an accuracy assessment of the
inverse kinematic solution by
evaluating how frequently solutions to the inverse kinematics problem were obtained by
\Cref{alg:inverse}, as well as evaluating the accuracy of those solutions.
The accuracy assessment was conducted using the following procedure:
\begin{enumerate}
  \item Randomly assign the set of $\theta_1,\dots,\theta_6$ within the feasible region of myCobot.
  \item Substitute the $\theta_1,\dots,\theta_6$ obtained in Step 1 into $A$ in
  \cref{eq:transformation-matrix} to determine the position and the orientation of the end-effector (solving the forward kinematics problem).
  \item Apply \Cref{alg:inverse} to the position and the orientation of the end-effector obtained in Step 2 (solving the inverse kinematics problem).
  \item Evaluate the number of times the solutions
  $\bm{\theta}_i=(\theta_{i,1},\dots,\theta_{i,6})$ obtained in Step 3
  match the $\theta_i$ given in Step 1, as well as the average number of different solutions obtained.
\end{enumerate}

\Cref{tab:accuracy-1,tab:accuracy-2} show the results.
For each test, 1000 samples of joint angles within the feasible region were randomly assigned, totaling 10000 samples.
In the table, each column shows the following:
``Avg.\ Time'' shows the average time it takes to solve the inverse kinematics problem in seconds\footnote{Note that the computing time was rounded to the 
nearest millisecond.},
``\# Success'' shows the number of cases in which the given joint angles were obtained%
\footnote{The criterion for whether the inverse kinematic problem was successfully solved is an absolute error of less than
$0.1$ [mm] between the position of the end-effector obtained from the solution of the inverse kinematics problem and the position of the given end-effector. 
This is a reasonable criterion since the repeatability accuracy of myCobot is $\pm 0.5$ [mm].},
``\# Diff. Soln.'' shows the number of cases in which solutions are different from the given one were obtained,
and ``Avg. \# Soln.''  the average number of solutions obtained.
The bottom row shows the average of each column.

\begin{table}[t]
  \centering
  \caption{Accuracy of the inverse kinematic solution \emph{without} pre-computation of the Gr\"obner basis (see \Cref{sec:accuracy-inverse})}
  \label{tab:accuracy-1}
  \begin{tabular}{c|cccc}
    \hline
    Test & Avg.\ Time [s] & \# Success & \# Diff. Soln. & Avg. \# Soln. \\
    \hline
    1 & 0.502 & 998 & 5 & 6.139  \\ 
    2 & 0.535 & 994 & 5 & 6.114  \\ 
    3 & 0.504 & 993 & 2 & 6.183  \\ 
    4 & 0.497 & 999 & 2 & 6.035  \\ 
    5 & 0.474 & 996 & 2 & 6.140  \\ 
    6 & 0.552 & 998 & 2 & 6.052  \\ 
    7 & 0.490 & 997 & 3 & 6.114  \\ 
    8 & 0.492 & 998 & 4 & 6.019  \\ 
    9 & 0.479 & 996 & 8 & 5.966  \\ 
    10 & 0.508 & 995 & 4 & 6.032  \\ 
    \hline
    Avg. & 0.503 & 996.4 & 3.7 & 6.0794  \\ 
    \hline
  \end{tabular}
\end{table}
\begin{table}[t]
  \centering
  \caption{Accuracy of the inverse kinematic solution \emph{with} pre-computation of the Gr\"obner basis (see \Cref{sec:accuracy-inverse})}
  \label{tab:accuracy-2}
  \begin{tabular}{c|c@{\hspace{14pt}}c@{\hspace{14pt}}c@{\hspace{14pt}}c}
    \hline
    Test & Avg.\ Time [s] & \# Success & \# Diff. Soln. & Avg. \# Soln. \\
    \hline
    1 & 0.112 & 999 & 2 & 6.056  \\ 
    2 & 0.097 & 993 & 0 & 6.122  \\ 
    3 & 0.097 & 999 & 2 & 6.038  \\ 
    4 & 0.097 & 995 & 1 & 6.062  \\ 
    5 & 0.097 & 999 & 2 & 5.974  \\ 
    6 & 0.097 & 997 & 4 & 5.944  \\ 
    7 & 0.097 & 997 & 0 & 6.112  \\ 
    8 & 0.097 & 998 & 3 & 6.126  \\ 
    9 & 0.097 & 998 & 2 & 6.112  \\ 
    10 & 0.097 & 997 & 2 & 6.086  \\ \hline 
    Avg. & 0.099 & 997.2 & 1.8 & 6.0632  \\ \hline 
  \end{tabular}
\end{table}

In the experiments shown in \Cref{tab:accuracy-1}, the pre-computation of the Gr\"obner basis described in \Cref{sec:optimization} was not performed, whereas in the experiments shown in
\Cref{tab:accuracy-2}, this pre-computation was carried out.
From both tables, we see that the pre-computation of the Gr\"obner basis reduces the computation time for solving the inverse kinematics problems to approximately one-fourth to one-fifth of the original time.
Furthermore, throughout the experiments, the success rate of solving the inverse kinematics problems has improved, and the number of cases in which the obtained solutions differ from the given original joint configurations has decreased.

In the following, we present our observations based on the results shown in \Cref{tab:accuracy-2}.
The table shows that \Cref{alg:inverse} was able to calculate approximately 99.7 \% of solutions to the inverse kinematic problems%
\footnote{For the remaining approximately 0.4\% of the inverse kinematic problems,
the position of the end-effector was not successfully obtained.
The cause may be related to numerical instability in solving the system of polynomial equations, but a detailed investigation of the cause 
is a future issue.}.
However, approximately 0.1 \% of those solutions differed from the initially given solutions.
(Note that, before optimization, the percentage of solutions that differed from the initially given solution was about 0.3 \%.)
This is likely due to errors arising from the numerical computation of trigonometric values in
Risa/Asir, and the propagation of the errors during the backward substitution of $x,y,z$, and each $\theta_i$.

We also see that the number of different solutions for each inverse kinematics problem averaged
about 6 per sample.
However, as mentioned earlier, there are cases where the solutions that should have been obtained
were not, so this average value is only for reference.

In the experiment, the initial configuration of each joint $\theta_1,\dots,\theta_6$ is randomly assigned in Step 1.
This is to guarantee the set of feasible positions and orientations for the end-effector.
As we have mentioned above, in myCobot, it may not be possible to independently determine the
position and the orientation of the end-effector since myCobot does not have three consecutive joints with intersecting rotational axes.
To verify this, we conducted another experiment by randomly assigning the position and the orientation of the end-effector and applying \Cref{alg:inverse} to solve the inverse kinematics problem.
\Cref{tab:random} shows the results of the experiment.
For each test, 1000 samples of the position and the orientation were randomly assigned, totaling 10000 samples.
In the table, the column ``\# Success'' shows the same value as \Cref{tab:accuracy-2}.
The result shows that only about 37.3 \% of solutions can be obtained when the position and orientation of the end-effector are randomly assigned.
This result suggests that there may be areas that cannot be reached by the end-effector
in a particular orientation, even within the operating range of the link, due to the structure of the manipulator or other reasons.

\begin{table}[t]
  \centering
  \caption{Results of the inverse kinematics computation with randomly given position and orientation}
  \label{tab:random}
  \begin{tabular}{c|cccccccccc|c}
    \hline
    Test & 1 & 2 & 3 & 4 & 5 & 6 & 7 & 8 & 9 & 10 & Avg.\\
    \hline
    \# Success & 393 & 371 & 374 & 355 & 336 & 386 & 360 & 393 & 355 & 406 & 372.9\\
    \hline
  \end{tabular}
\end{table}
\subsection{Accuracy of the Forward Kinematic Solution}
\label{sec:accuracy-forward}

Next, we conducted an accuracy assessment of the forward kinematic solution by
evaluating the discrepancy between the desired position and orientation of the end-effector
for the inverse kinematics problem, and the position and orientation
obtained from \Cref{alg:inverse} by the following procedure.
\begin{enumerate}
  \item Randomly assign the set of $\theta_1,\dots,\theta_6$ within the feasible region of myCobot.
  \item Substitute the $\theta_1,\dots,\theta_6$ obtained in Step 1 into $A$ in
  \cref{eq:transformation-matrix} to determine the position and orientation of the end-effector (solving the forward kinematics problem).
  \item Apply \Cref{alg:inverse} to the position and the orientation of the end-effector obtained in Step 2 (solving the inverse kinematics problem).
  \item Substitute the solution $\theta_i$ obtained in Step 3 into $A$ in
  \cref{eq:transformation-matrix}  to obtain the position and the orientation of the end-effector.
  \item Evaluate the errors of the position and orientation obtained in Step 4 from those obtained in Step 2.
\end{enumerate}

As the function used for error evaluation (error evaluation function),
we employ the error rate (relative error) $\varepsilon$, defined as
$|(x-x')/x|$,
where $x$ is the theoretical value and $x'$ is the measured value.

Let $E'_i$ be the error evaluation function for one of the solutions to the inverse kinematics problem obtained by \Cref{alg:inverse}.
$E'i$ is obtained from the average error rate of six input values: three related to orientation  ($\alpha, \beta, \gamma$) and three related to position ($p_1, p_2, p_3$). Therefore, if the error rates for each input value are $\varepsilon_\alpha, \varepsilon_\beta, \varepsilon_\gamma, \varepsilon_{p_1}, \varepsilon_{p_2}, \varepsilon_{p_3}$, $E'_i$ is defined as
$E'_i = (\varepsilon_\alpha+\varepsilon_\beta+\varepsilon_\gamma+\varepsilon_{p_1}+\varepsilon_{p_2}+\varepsilon_{p_3})/6$.
\Cref{tab:accuracy-2} shows that there is an average of about six solutions to the inverse kinematics problem per sample.
Therefore,
for the error evaluation functions $E'_i$ calculated for each solution,
the final error evaluation function $E$ corresponding to a sample point is defined as the average value of $E'_i$ $(i=1,\dots,6)$, expressed as
\begin{equation}
  \label{eq:error}
  E = \frac{1}{\# A}\sum_{i \in \{1,\dots,\# A\}} E'_i,
\end{equation}
where $A$ represents the set of solutions to the inverse kinematics problem obtained from
\Cref{alg:inverse}, and $\# A$ denotes the number of elements in $A$.

\begin{figure}[t]
  \centering
  \includegraphics[width=\linewidth,scale=1.2]{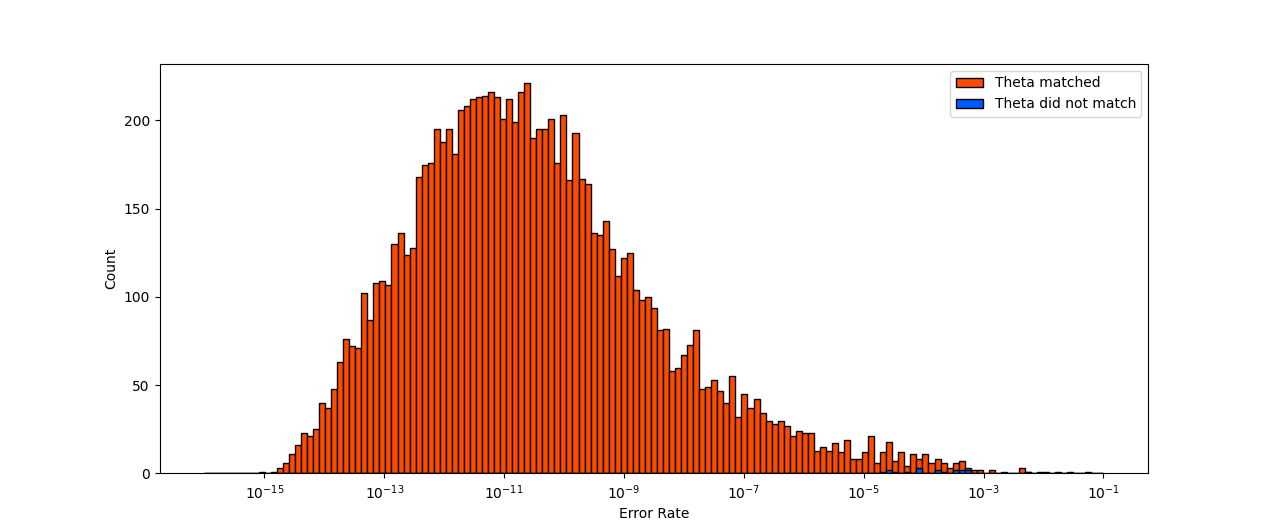}
  \caption{Distribution of error rates of the solutions}
  \label{fig:error-2}
\end{figure}

\Cref{fig:error-2} shows the results of the error evaluation.
Note that the randomly generated samples here are
the same as those used in \Cref{sec:accuracy-inverse}.
In the figure, the samples for which solutions to the inverse kinematics problem were obtained are shown in bright color, and the samples where the initially given $\theta_1,\dots,\theta_6$ did not match the output of \Cref{alg:inverse} are shown in dark color.
From \Cref{fig:error-2}, we see that approximately 95 \% of the results have an error evaluation
function $E$ of less than $10^{-6}$.
Considering that the theoretical range of the end-effector's position ($p_1,p_2,p_3$) is approximately $-200$ [mm] to $400$ [mm], the positional error 
is about $10^{-4}$ [mm].
Since the reproducibility of myCobot is $\pm 0.5$ [mm],
we see that the output of \Cref{alg:inverse} has sufficient accuracy.
On the other hand, for the samples where the initially given
$\theta_1,\dots,\theta_6$ did not match the output of \Cref{alg:inverse},
most of them had an error evaluation function $E$ of $10^{-5}$ or more.
Considering the reproducibility of myCobot, the accuracy of the output of \Cref{alg:inverse}
cannot be considered good for these samples.

From both the evaluation of the errors in the forward and inverse kinematic solutions,
investigating the causes, and proposing solutions for the phenomenon where the output of
\Cref{alg:inverse} does not match the initially given $\theta_1,\dots,\theta_6$, remains future work.

\section{Concluding Remarks}
\label{sec:conclusion}

In this paper, we have proposed a solution to the inverse kinematics problem of the 6-DOF manipulator
myCobot.
The proposed method determines the intersection of the rotation axes for two consecutive joints where the axes intersect.
Then, by using the coordinates of the intersection, it derives the rotation angles of each joint.
To calculate the rotation axes, we have derived a system of polynomial equations with the coordinates $x,y,z$ of the rotation axes as variables and the position and orientation of the
end-effector as parameters within the coefficients.
When solving the system of equations, we have computed the CGS of the polynomial system.
By avoiding repeated calculations of the Gr\"obner basis for different positions and orientations, we aim to improve the overall efficiency of the method.
In computer experiments, we have demonstrated that the inverse kinematics problem is solved with sufficient accuracy for practical use compared to the size of myCobot.

The challenges of this study are as follows.

First, we need to identify the causes of the inverse kinematics problems that failed to produce correct solutions and to develop appropriate countermeasures. In addition, for cases where the obtained solutions differ from the originally given joint configurations, it is also necessary to investigate the underlying causes and consider suitable remedies.

Second, since the calculation of the CGS is not complete, there are segments that are not covered. One countermeasure is to narrow the parameter space. Due to the structure of myCobot,
it may be possible to narrow the parameter space required for solving the inverse kinematics problem.
Another countermeasure is to utilize new algorithms or implementations for the CGS
\cite{nab2024,wad-nab2024}.

Third, the calculation of trajectory planning using the proposed method.
The trajectory planning problem is an extension of the inverse kinematics problem, where the latter involves solving the problem at a single point, while the former requires solving it over an entire given trajectory.
In trajectory planning, when there are multiple solutions to the inverse kinematics problem at a certain point on the trajectory, the issue is which solution to choose to move the joints.
We have proposed a method that reduces this problem to the shortest path problem in a graph and uses Dijkstra's algorithm to find the optimal combination of solutions to the inverse kinematic problem \cite{shi-oka-ter-mik2024}.
We are currently developing a trajectory planning method that combines the proposed method with this approach \cite{oka-ter-mik2025b}.

\section*{Acknowledgments}
  The authors would like to thank the anonymous reviewers for their helpful comments.
  This research is partially supported by JKA and its promotion funds from KEIRIN RACE.
  In addition, this work was partially supported by the Research Institute for Mathematical
  Sciences, an International Joint Usage/Research Center located at Kyoto
  University.

\bibliographystyle{splncs04}
\bibliography{casc-2025-okazaki-terui-mikawa}

\end{document}